\documentclass{article}

\usepackage{arxiv}
\usepackage[usestackEOL]{stackengine}
\usepackage[utf8]{inputenc} 
\usepackage[T1]{fontenc}    
\usepackage{hyperref}       
\usepackage{url}            
\usepackage{booktabs}       
\usepackage{amsfonts}       
\usepackage{nicefrac}       
\usepackage{microtype}      

\usepackage{array}
\usepackage{graphicx}
\usepackage{natbib}
\usepackage{doi}

\makeatletter
\renewcommand\scriptsize{\@setfontsize\scriptsize{7}{8}}
\renewcommand\tiny{\@setfontsize\tiny{5}{6}}
\renewcommand\small{\@setfontsize\small{12}{14}}
\renewcommand\normalsize{\@setfontsize\normalsize{12}{14}}
\renewcommand\large{\@setfontsize\large{14}{16}}
\renewcommand\Large{\@setfontsize\Large{14}{16}}
\renewcommand\LARGE{\@setfontsize\LARGE{14.4}{18}}
\renewcommand\huge{\@setfontsize\huge{20.74}{30}}
\renewcommand\Huge{\@setfontsize\Huge{24}{36}}
\makeatother

\title{Analysis of the use of color and its emotional relationship in visual creations based on experiences during the context of the COVID-19 pandemic}

\date{} 					

\author{ 	\href{https://orcid.org/0000-0002-9017-3196}{\includegraphics[scale=0.06]{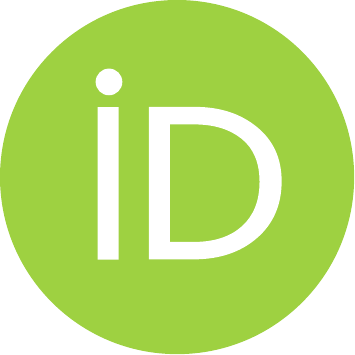}\hspace{1mm}Gonz\'alez-Mart\'in, C\'esar } \\
	Faculty of Education Science \\
	Universidad de Córdoba,\\
	C\'ordoba, Spain\\\thanks{corresponding author: \texttt{cgonzalez@unizar.es}}  \\
	\And
	\href{https://orcid.org/0000-0002-5389-7590}{\includegraphics[scale=0.06]{orcid.pdf}\hspace{1mm}Carrasco, Miguel} \\
	Facultad de Ingeniería y Ciencias\\
	Universidad Adolfo Ibañez\\
	Santiago, Chile\\
	\And
	\hspace{1mm}Oviedo, German\\
	Facultad de Ingeniería y Ciencias\\
	Universidad Adolfo Ibañez\\
	Santiago, Chile\\
}

\renewcommand{\shorttitle}{Analysis of the use of color and its emotional relationship in visual creations based on experiences during the context of the COVID-19 pandemic}

\hypersetup{
pdftitle={Analysis of the use of color and its emotional relationship in visual creations based on experiences during the context of the COVID-19 pandemic},
pdfsubject={},
pdfauthor={Gonzalez-Mart\'in, C., Carrasco, M.},
pdfkeywords={color,emotion,deep learning,covid-19,pandemic},
}

\begin{document}
\strutlongstacks{T}
\maketitle

\begin{abstract}
Color is a complex communicative element that helps us understand and evaluate our environment. At the level of artistic creation, this component influences both the formal aspects of the composition and the symbolic weight, directly affecting the construction and transmission of the message that you want to communicate, creating a specific emotional reaction. During the COVID-19 pandemic, people generated countless images transmitting this event's subjective experiences. Using the repository of images created in the Instagram account CAM (The COVID Art Museum), we propose a methodology to understand the use of color and its emotional relationship in this context. The process considers two stages in parallel that are then combined. First, emotions are extracted and classified from the CAM dataset images through a convolutional neural network. Second, we extract the colors and their harmonies through a clustering process. Once both processes are completed, we combine the results generating an expanded discussion on the usage of color, harmonies, and emotion. The results indicate that warm colors are prevalent in the sample, with a preference for analog compositions over complementary ones. The relationship between emotions and these compositions shows a trend in positive emotions, reinforced by the results of the algorithm a priori and the emotional relationship analysis of the attributes of color (hue, chroma, and lighting).

\end{abstract}

\keywords{color \and emotion \and deep learning \and covid-19  \and pandemic}

\section{Introduction}
The pandemic caused by the coronavirus (COVID-19) generated a collective experience worldwide with limitations on our freedoms through confinement, maintenance of interpersonal distances, travel restrictions, and indoor capacity of small spaces. Also, introducing new elements such as masks, hydroalcoholic gels, or gloves, among other aspects.  These have transformed our daily lives and have become elements of representation to express and communicate the individual and collective experiences produced during the pandemic, generating an innumerable amount of visual material shared through several channels.

Social networks have seen their use and presence increase during the pandemic, becoming one of the most used sources to be informed, entertained, and share experiences, especially those based on visual content such as Instagram and TikTok.

Online initiatives emerged, such as the virtual museum called COVID Art Museum (CAM) on Instagram (@covidartmuseum), which is curating a collection of visual works made by internet users as a repository of user-generated contents (UGC), where the leitmotiv is the pandemic and its effects. In this way, these digital creations become a source of information to understand and analyze the subjective experiences lived during the pandemic and identify common patterns among users in representation strategies to communicate their experiences through visual language as it has been a global event. 

Visual creation involves making decisions about using representative elements, which affect the formal aspects of the image (low-level) and its significance and symbolic charge (high-level). Color acts on both levels within these decisions and can express abstract concepts~\citep{schloss_blue_2020}.

This research aims to analyze the use of color on the images that make up the CAM collection, to understand at a formal and emotional level the use of color to build visual representations of the subjective experiences lived during the pandemic through a methodology of quantitative visual analysis based on deep learning tools and combined analysis between emotion and color. 

Through the proposed methodology, it is possible to detect common patterns in users' creations to understand if there are recurrent emotional discourses through color attributes.

\subsection{About the color}
We talk about additive colors (generated by light) and subtractive (by pigments) depending on the source. Within these, there are primary colors – red, green, and blue (RGB) in the case of additives and cyan, magenta, and yellow in pigment colors – which, through their combinations, will result in secondary and tertiary colors. White, black, and gray colors are considered achromatic and regulate the tones' saturation and lighting, which helps build the aesthetic experience and symbolism. However, these can work independently, equally creating an aesthetic and an emotional message.

Thus, color is a semantic indicator that affects the image's composition, influencing the visual weights of balance and imbalance through the manipulation of the hue (color), through its saturation (chroma) and luminosity (value), and its combinations with other colors, generating contrasts, rhythms, and harmonies. For example, the value alteration affects the psychology and perception of color temperature~\citep{eiseman_complete_2017}. The alteration of color attributes helps to build the aesthetic experience, knowledge of abstract semantics, communication \citep{jacobson_color_1996,kauppinen-raisanen_using_2018,wei-ning_image_2006} and mark preferences over colors \citep{jalali_palette_2016,ou_study_2004-2}. 

There are models to describe the relationships between colors \citep{amirshahi_evaluating_2014}. Eiseman \citeyear{eiseman_complete_2017} points out the most conventional ones: monotones, that there is only one neutral color (off-whites, beige, grays, and taupe are considered neutral). The monochromatic, where a hue and its chromatic variables are used. Analogs or adjacent tones are close to each other in the chromatic circle and share a common color (usually a primary) in their composition, creating the most harmonies \citep{white_color_2021}. The complementary tones, which are directly opposite in the chromatic circle, create a simultaneous contrast that visually vibrates both tones (see more in \citep{pridmore_complementary_2021} ). The split-complementary are two adjacent colors combined with a complimentary in the chromatic circle. Triads are three colors equidistant at an angle, and finally, tetrads are two pairs of complementary colors. Thus, through color combinations, we can accentuate the characteristics of the colors that compose it and create more or less harmonious, balanced, and rhythmic compositions.

In addition to hue, chroma, and value, other factors sometimes alter color perception \citep{mogi_color_2021}.  Elliot \citeyear{elliot_color_2015} points out some as "(...) viewing distance and angle, amount and type of ambient light, and presence of other colors in the immediate background and general environmental surround" (p.5).

\subsection{Significance, symbology, and emotion in relation to color}
Research on color is very broad and extends to different areas of knowledge, such as marketing and advertising \citep{gorn_effects_1997,meyers-levy_understanding_1995}, design in different fields \citep{baptista_effects_2021,cyr_colour_2010}, consumer behavior \citep{hsieh_colors_2018,jalali_palette_2016}, psychology \citep{elliot_color_2015,reece_instagram_2017}, tourism \citep{yu_coloring_2020,yu_color_2021}, in therapeutic treatments \citep{luo_color_2015,whitfield_arcane_2015}, and of course, art. His research goes back decades, approached from its perceptual characteristics at the psychophysical level \citep{berthier_relativity_2021,hsu_chromaticity_2012,moutoussis_physiology_2015,stasenko_when_2014,xiong_detection_2021}, and its link with emotional processes \citep{elliot_color_2015,elliot_color_2007,luo_color_2015}.

Color affects the perceptual and cognitive level, which ''(...) in the literature of psychology, it is generally considered that sensation and perception refer to the immediate mapping of objects or events of the real world into the brain, while cognitive represents subsequent higher-order processes of semantic and verbal classification of the perceptions ''\citep[p.411]{gao_investigation_2006}.  Color is a source of knowledge and becomes a mechanism to evaluate situations consciously, semi-consciously, and unconsciously \citep{kauppinen-raisanen_using_2018}, determining our acceptance or rejection of the phenomena that occur in our environment, which implies an influence on our behavior \citep{chan_color_2021,mehta_blue_2009,suk_emotional_2010,valdez_effects_1994}.

The complexity of the study of color lies in its characteristics as an element that can function as a sign and as an object in itself, transmitting a meaning without being associated with an object \citep{caivano_color_1998}.  For example, dark colors produce sensations of being heavier than luminous ones, but a white stone and a black stone can physically weigh the same. When color is associated with another, the syntax of color (sign + sign) occurs a visual relationship that develops the aesthetic experience and a more complex conceptual message. In these relationships, we face color most regularly.

At the semantic level, color can function as an icon through a resemblance to an object —for example, orange is associated with fire. The color as an index is an indication, as is the example of the color of the fruit in its ripening, or the ''white on detergent packaging is a typical indexical sign that conveys the associative meaning of purity and cleanliness, while yellow on packaging containing vitamin C suggests the indexical meanings of energy through its resemblance with the sun,  the main source of energy''\citep{kauppinen-raisanen_using_2018}.  As a symbol, color is something we learn, a convention \citep{schloss_blue_2020}, such as red and a traffic light, means to stop driving, or certain colors on a flag represent a specific community. All this is learning.

Our relationship with colors and their association with events, objects, concepts, emotions, etc. \citep{retter_early_2021}, is assumed through the repetition of learnings based on experiences \citep{elliot_color_2007}, with which individual knowledge is acquired, and that is why there are people who differ when it comes to associating an adjective with a color \citep{manav_color-emotion_2007}.  For example, the preference over one color over another may be determined by the association between color and object according to a color concept network \citep{schloss_ecological_2017} or The ecological valence theory proposed by Palmer and Schloss~\citeyear{palmer_ecological_2010}.  But, ''The reasons for color preference are fluid and diverse.  Some peoples' preferences may be governed by object associations, others by basic psychophysical dimensions, others by biological components of color vision'' \citep[p.1026]{taylor_color_2013}.  For example, due to their biological conditions, dichromats prefer yellow colors over blues, and trichromats prefer the latter \citep{schloss_color_2015}.

If personal peculiarities influence the perception and significance of color, and learning is personalized, color cannot be shared universally \citep{schloss_ecological_2015,taylor_color_2013}. Thus, there is no evidence of universality on color preference from a gender perspective in research, detecting some differences between sexes \citep{gilbert_color_2016,hurlbert_biological_2007,kim_instagram_2019,ou_study_2004,rosenbloom_color_2006,taylor_relationship_2012}.  Thus, Ou \citeyear{ou_study_2004-1} points out that ''female observers tended to associate "like" with the color pairs that were "light," "relaxed," "feminine," or "soft," whereas this association did not occur for male observers'' (p.298).  For his part, Valdez and Albert \citeyear{valdez_effects_1994} finds that women are more sensitive to luminosity and saturation, although he recognizes that the emotional reaction is similar. 

Another characteristic that marks differences in color preferences and emotions is age \citep{gilbert_color_2016,knoblauch_variation_2001,manav_color-emotion_2007,ou_age_2012,palmer_ecological_2010,terwogt_colors_1995}, which explains that ''(...) color preferences can and do vary systematically over time'' \citep[p.95]{schloss_ecological_2017}.  On-time, Palmer and Schloss \citeyear{palmer_ecological_2010} point out that preferences over colors are kept or changed, based on the experiences with the object, or the change ''(...) such as boredom, new physical or social circumstances, or fashion trends, change the dynamics of aesthetic response, as indeed they inevitably do'' (p.8881).

If we look at cultures and geographical situations, Ou et al. \citeyear{ou_study_2004} point out differences in emotions concerning color. However, in a subsequent study, no significant differences are pointed out in the combination of colors in relation to emotions \citep{ou_study_2004-1}. For its part, Xin~\citeyear{xin_cross-regional_2004,xin_cross-regional_2004-1}
 shows specific differences between regions about color attributes. Comparing industrialized and non-industrialized societies, Taylor \citeyear{taylor_color_2013} finds few coincidences in color preferences.  Elliot \citeyear{elliot_color_2007} points out that particular meanings and effects produced by color are associated with cultures. 

On the contrary, Gao et al. \citeyear{gao_investigation_2006} find uniformity between perception and emotion between cultures, thus eliminating experiential learning. Jonauskaite \citeyear{jonauskaite_machine_2019} brings significant coincidences between members of the same cultural group to recognize emotions and find universalities between groups.  For its part, Carruthers \citeyear{carruthers_manchester_2010} does not find significant differences between healthy patients or patients with mood issues.  In the face of theories of experiential learning, this author suggests that associative learning between color and emotion does not necessarily imply learning based on direct experience with color but occurs through conceptual learning and language \citep{jonauskaite_colour-emotion_2021}. Hence, people with visual difficulties can be consistent with people without visual problems. For this reason, the author points out that the color-emotion association is universal: ''Moreover, there is a global similarity in how specific emotion concepts are associated with specific color terms, although these universal associations are modulated by geographic and linguistic factors''\citep[p.11]{jonauskaite_machine_2019}. 

Jahanian \citeyear{jahanian_colors_2017} also finds a high agreement level between linguistic association and color in different cities, genders, and age groups, explaining the conventions discussed above since emotions must be translated into language and related to colors.  Thus, we talk about red-passion, warm and cold colors, light-heavy, good-bad, comfortable, etc. \citep{gong_investigation_2017,schloss_blue_2020}, and sometimes they have no natural relationships. Hence, signs of universality can be found.

Schloss \citeyear{schloss_color_2018} proposes an interesting Color inference framework where three cognitive processes run that explain how conceptual inferences are made on the information contained in the colors that influence our judgment about the environment.

Different relationships have been demonstrated regarding the impact of color attributes to generate sensations and emotions. Valdez et al. \citeyear{valdez_effects_1994} showed that saturation and value have an important effect on emotions. Following Russell and Mehrabian \citeyear{russell_distinguishing_1974}, it finds a high saturation level generates more arousal. The luminous colors are more pleasure than arousal or dominance, affecting the activity and potency, while the dark ones are the opposite, with higher levels of dominance and arousal than pleasure. 

Along the same lines, Suk~\citeyear{suk_emotional_2010}, Manav~\citeyear{manav_color-emotion_2007}, and  Schloss~\citeyear{schloss_blue_2020} agree that the most influential factors for an emotional reaction are value and chroma \citep{manav_color-emotion_2007}, but all depend on the context of the observer~\citep{xin_cross-regional_2004,xin_cross-regional_2004-1}. 

Gong \citeyear{gong_investigation_2017} and Ou, et al. \citeyear{ou_study_2004} point out differences in the influence of the attributes depending on the sensation you want to achieve. Thus the chroma is more decisive for the sensation in the color temperature (cold-warm), while the lighting produces the Heavy-light sensation. In some cases, such as the hard-soft feel, it is related to both attributes, chroma, and luminance~\citep{ou_study_2004}. Schloss's study \citeyear{schloss_blue_2020} shows that judgment about happiness and sadness is determined by the chroma and luminosity of color, noting that dark blue is happier than dark yellow. On the contrary, at the perceptual level and preference in color, the hue attribute seems to be more determinant than the value and chroma \citep{gong_investigation_2017}.

\subsection{Color and emotion: from computing to deep learning}
From a computational perspective, the analysis of emotions expressed on a digital image has achieved an important leap in the last decade \citep{chen_learning_2015,kim_deep_2013,ram_extrapolating_2020,yao_apse_2020}.  In the beginning, the primary strategies focused on the extraction of low-level characteristics such as color \citep{corridoni_image_1999}, texture or aesthetics \citep{datta_algorithmic_2008,joshi_aesthetics_2011,machajdik_affective_2010,marchesotti_assessing_2011}, based on the principles of art composition  \citep{zhao_exploring_2014}, or through the recognition and identification of the objects that make up an image,  seeking to connect the description of the elements with the emotion \citep{borth_large-scale_2013,vonikakis_emotion-based_2012,yanulevskaya_emotional_2008}.  Unfortunately, these methods have had limited results. They focus on the description of features, with the observations and analysis performed by a human, a process known as hand-crafted features \citep{zheng_feature_2018}.

The extraction of emotion in digital images is more complex than other types of operations since it is subjective to the evaluator and can be influenced by different factors such as color, shapes, the appearance of lines, and the emotional bias of the observer himself \citep{hanjalic_extracting_2006}. This makes recognizing emotions a complex problem since they are not presented in their pure form on an image. The combination of different emotions on the same image is often present \citep{yang_joint_2017,zhao_exploring_2014} making its evaluation subjective and difficult to determine empirically.  To attach the images with a particular emotion, it was necessary to use a categorical model from psychology, which is based on the categories of emotional states (CES) and in the space of dimensional emotion (DES). The CES model considers a limited set of emotions such as fear, disgust, sadness, happiness, among others \citep{mikels_emotional_2005,wei-ning_image_2006}. Instead, the DES model combines three variables: valence, arousal, dominance \citep{plutchik_measurement_1989,schlosberg_three_1954}. In them, arousal describes the intensity of emotion; valence is related to the type of emotion; and dominance with submission and control. Usually, the latter is not used since it does not have relevant information on the images. For this reason, the valence-arousal scheme is generally used \citep{xu_hierarchical_2013}. 

Despite these limitations, a new approach to solving the problem of analysis and extraction of emotions in different types of images has emerged thanks to the development driven by Convolutional Neural Networks (CNN) \citep{krizhevsky_imagenet_2017}.  CNN allow you to extract patterns automatically without requiring manual extraction work.  CNN manages to extract and generate discriminative characteristics through a supervised learning process where the categories of the problem are previously known. In this way, CNN manages to separate highly non-linear and correlated characteristics on different sources of information.  At present,  the range of applications involving CNN is vast and diverse, achieving the extraction of patterns on various problems and domains \citep{liu_application_2020,poterek_deep_2020,reno_combined_2020,Wang_2018,Westlake_proceedings_2016,zhuang_online_2021}. 

The first advances in detecting emotions through CNN correspond to the work developed by  Kim et al. \citeyear{kim_deep_2013} and Razavian et al. \citeyear{razavian_cnn_2014}. The authors used the network proposed by Krizhevsky \citeyear{krizhevsky_imagenet_2017} with some modifications both in the increase of data or through incorporating other sources of information (labels or audio).  Therefore, the research focused strongly on face recognition, an active research area in recent years. In this category are the works proposed by Kahou et al. \citeyear{kahou_emonets_2015}, Kollias et al.  \citeyear{kollias_interweaving_2015} and Wei et al.  \citeyear{wei_new_2017}.  Unfortunately, these solutions lose generality as they are strongly focused on the primary detection of the face without considering other aspects that make up the image. 

An essential key to interpreting emotion lies in analyzing the objects present in the images. However, this is not the only information available.  In this regard, Kim et al. \citeyear{kim_building_2017} and Priya et al.\citeyear{priya_affective_2020} aim to integrate both information from objects and the background of images and low-level features.  Although color is a low-level characteristic, it has an important relationship with the object and the image's meaning.  On the other hand, Rao et al.~\citeyear{rao_learning_2018} propose the combination of multiple CNN that consider the detection at a high level of the semantics of objects, such as colors, textures, and even aesthetics \citep{lu_rapid_2014}. Although CNN networks initially did not use these elements, their incorporation has improved the performance at a global level in the detection of emotion.

Although color begins to be relevant in the analysis, it is essential to understand that color can generate different emotions according to our cultural identity \citep{jonauskaite_machine_2019}. It is important to understand that color is not a static unit; it can be transferred and thereby affect the emotion generated in the observer \citep{elliot_color_2015}.  One of the first works in this area was proposed by He et al. \citeyear{he_image_2015} which offered a tool to modify the color according to the target emotion. Other works have deepened this analysis either through the combination of different CNN \citep{liu_emotional_2018} or by employing a relationship between color and texture, improving the relationship of color harmony \citep{liu_texture-aware_2018}.

In recent years, the analysis of color and how it affects the interpretation of emotion has begun to arouse greater interest from the scientific world \citep{elliot_color_2015}.  For example, Ram et al.  \citeyear{ram_extrapolating_2020} studied the relationship between color and emotions through an extensive questionnaire applied to more than 900 subjects.  It is interesting to note that the results may vary more by the observer's age than by gender. However, not all color combinations generate a correlation with emotion.  In this sense, Takada et al. \citeyear{takada_color-grayscale-pair_2021} analyze the meaning of emotion when analyzing color and grayscale images.  It is interesting to note that color images generate a more significant effect (positive or negative) on the observer. 

\subsection{Affective computing}
Although the first theoretical approaches to affective computing have existed for more than two decades \citep{minsky_society_1986,picard_affective_1995} in recent years, it has been possible to advance in the understanding of images and how they induce different emotions in the observer \citep{zhao_affective_2020} mainly due to the enormous advance in computational techniques. Zhao et al.  \citeyear{zhao_affective_2021} call this task ''Analysis of the Content of Affective Image'' and use as input the development of emotional models based on affective image datasets and quantitative methods to carry out this task.

As we discussed earlier, different emotional models derived from psychology have been used as a theoretical framework to measure possible perceived emotions.  Zhao et al.  \citeyear{zhao_affective_2020} divide these models into two: Categorical Emotion States, where their basic manifestation would be ''sentiment'' (negative, positive, and sometimes neutral categories); and Dimensional Emotion Spaces where emotions are distributed in a continuous space, such as the VAD model of valence-arousal-dominance.  From a computational and machine learning perspective, it is necessary to have images as representative labels of some of the emotional models mentioned above \citep{cacioppo_emotion_2007}. A dataset widely used in psychology is IAPS \citep{lang_international_1999} with 1,182 images labeled in the VAD model and which has also been taken to a DES model in its IAPSa format \citep{mikels_emotional_2005} with 390 images. Today, with the presence of social networks, it is possible to see datasets with a more significant number of images, such as IESN \citep{zhao_predicting_2016} with more than 1 million images tagged in both VAD and Mikel's CES model and T4SA with almost 1.5 million images categorized under the ''Sentiment'' model.

In general, datasets are composed of digital photographs \citep{zhao_affective_2021}. However, there are also datasets focused on the domain of artistic images. In this line, the datasets of ArtPhoto \citep{machajdik_affective_2010} , MART \citep{alameda-pineda_recognizing_2016}, devArt \citep{alameda-pineda_recognizing_2016}, WikiArt \citep{saleh_large-scale_2015} and ArtEmis \citep{achlioptas_artemis_2021}, all in Categorical Spaces of Emotion. Both WikiArt and ArtEmis share that the WikiArt project is used as an image source (Table \ref{tab:table_1}).

\begin{table}
	\caption{Emotion dataset properties}
	\centering
	\scalebox{0.9}{
	\begin{tabular}{p{2cm}p{4cm}p{2cm}lp{2cm}p{4cm}}
Dataset  & Reference& Image number & Type & Emotion Model & Tags\\
\toprule
ArtPhoto	&\citep{machajdik_affective_2010}			&806	& Artistic	&Mikels 		&1 DEC category per image\\
MART	&\citep{alameda-pineda_recognizing_2016}	& 500	&Abstract&Sentiment	&1 DEC category per image\\
devArt	&\citep{alameda-pineda_recognizing_2016}	&500	&Abstract&Sentiment	&1 DEC category per image\\
WikiArt Emotions	&\citep{saleh_large-scale_2015}		&4105	&Artistic	&Custom DEC model&DEC Relative Weight per Image\\
Artemis	&\citep{achlioptas_artemis_2021}			&80000	&Artistic	&Mikels	&DEC Relative Weight per Image\\
\toprule
	\end{tabular}}
	\label{tab:table_1}
\end{table}

\section{ Methodology}
The proposed methodology focuses on understanding at a formal and emotional level the use of color in the visual representations of the subjective experiences lived during the pandemic, using the collection of images of the Covid Art Museum (CAM) of Instagram. For this, an analysis is proposed that includes three stages: I) extraction of emotions based on deep learning techniques, II) extraction of color and its harmonies, and III) fusion of emotion and color information.  Below, we detail each of them.

\begin{figure}
	\centering
	\includegraphics[width=\textwidth]{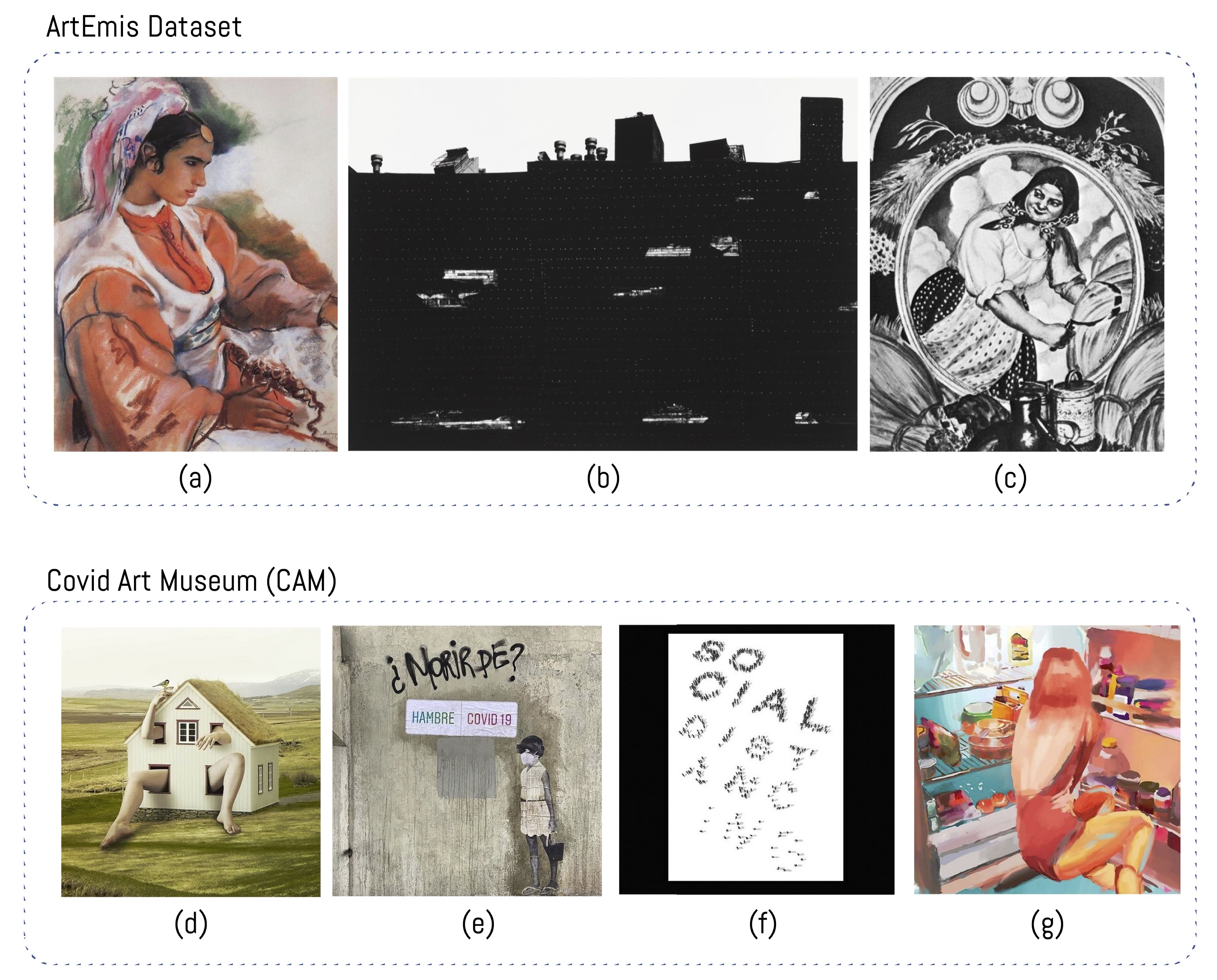}
	\caption{ Sample images from ArtEmis Dataset and Instagram's Covid Art Museum account (a) Young Moroccan, Zinaida Serebriakova (1932);  (b) New York I, Aaron Siskind (1951); (c) Harvester, Boris Kustodiev (1918),(d) @marciorodriguezphoto, (e) @Bix.rex.1 (f) Hojin Kang, @hojinkangdotcom (g) Pauz Peralta, Title: The Fridge, @notpauz}
	\label{fig:fig1}
\end{figure}

\begin{figure}
	\centering
	\includegraphics[width=\textwidth]{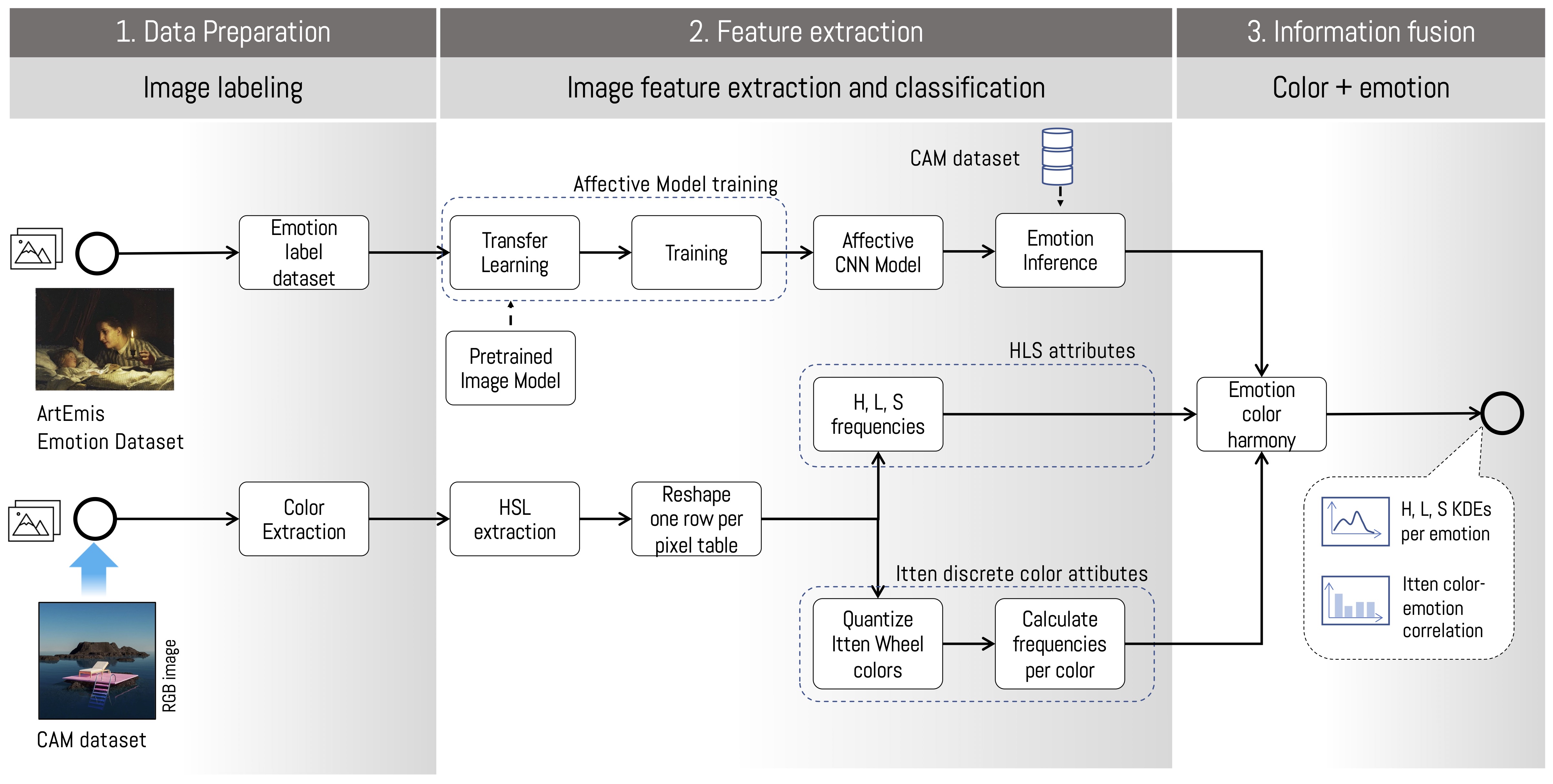}
	\caption{Process diagram of the proposed method for the analysis of emotion and color harmonies applied on the Covid Art Museum (CAM) dataset}
	\label{fig:fig2}
\end{figure}

\begin{description}
\item[ Step I) Extraction of Emotions:] As we have previously described, techniques based on deep learning have made enormous advances for extracting emotions in different types of images.  This research uses, in particular, the ArtEmis dataset  (Fig.\ref{fig:fig1})," which has the largest number of artistic images associated with a feeling (see Table~\ref{tab:table_1}).  The ArtEmis dataset consists of one or more files separated by CSV commas which contain a web address of an image available on the WikiArt platform. Each of them is associated with nine numerical attributes (one per emotion), which encode the emotions of people who observed and reported the emotions they felt during the experiments \citep{saleh_large-scale_2015}. The standard library for HTTP requests (via the Python 3.8 language) was used for download. In this way, a unique identifier related to the URL described in the CSV was used for each downloaded image. The resulting dataset corresponds to 78,004 images in conjunction with their identification data.

The training stage of the model consists of designing a CNN network that allows understanding emotions of a previously labeled dataset; in this case, on the ArtEmis dataset.  This allows the CNN network to extract underlying patterns in the data which describe emotions in digital images.  Subsequently, the learning acquired on ArtEmis is applied to the CAM dataset (Fig.\ref{fig:fig1}) through inference of the model.  The CAM dataset was extracted using a web scraping technique using Open Source Intelligence software \citep{hassan_open_2018}. This software allows the massive extraction of information through a command line. For this study, 1537 images were downloaded from the CAM in May 2021, corresponding to different types of digital art such as photographs, photomontages, digital paintings, and drawings.

The inference stage allows the extraction of emotions associated with a vector of probabilities for each of the images of the CAM.  This process is relevant since the inference is made on an unknown set in the CNN network training process.  Subsequently, the network generates a vector of emotions that will later be related to the color methodology.

\item[Step II) Color Analysis:] This stage uses color extraction to convert the RGB color model to the HSL (Hue-Saturation-Value) model.  The HSL model is considered intuitively more straightforward than the RGB model to explain the color mixing \citep{rhyne_applying_2017}.  This type of transformation allows an aesthetic analysis of color \citep{eiseman_complete_2017}.  The colors used by an artist depend on light, so it is appropriate that it is a dimension of the color model to be used for artistic analysis \citep{arnheim_art_1954}. Once this process is complete, each channel (HSL) is transformed into a list of independent values.  Then, two threads are performed in parallel.

\subitem[II.a] 	Estimation of frequencies through a histogram and subsequent estimation of a density function (KDE).  The density function makes it possible to estimate the probability of a given function. This is relevant when we will combine emotions within a particular distribution.

\subitem[II.b]	Color quantization based on the Itten color palette.  The colors of each image are quantized in relation to the Itten palette. Then, the composition of each group and the harmony relationships present in the palette are analyzed.  In particular, relationships of the type analogs, complementary, split-complementary, triads in the clustered groups previously, among others, are sought.  Thus, it is possible to inquire about the essence of colors in the image. This type of analysis allows to find the harmonies detected according to the frequencies of relationship, which is related to the composition and aesthetics, whether more or less harmonious, balanced, or rhythmic. For this, an angular measurement between the colors expressed in the Itten palette is used to determine the relationship between the colors that make up the color wheel. 

\item[Step III) Information fusion:] Once the emotion, color harmonies, and density functions (KDE) of each color are extracted, the information is correlated, allowing a descriptive level of greater depth by providing information both in form and background of artwork.  The proposed method makes it possible to associate a density function with emotion, and the same goes for the frequency of harmonies to colors.   The entire above process is described in detail in Fig.\ref{fig:fig2}.

\end{description}

\section{Results}
This section presents the results obtained by the proposed methodology and the discovery of emotional discourses through the use of color attributes. As we will review below, the analysis of color and emotions presented in the methodology uses two independent paths processed in combination.

\subsection{Color analysis}
\begin{figure}
	\centering
	\includegraphics[width=\textwidth]{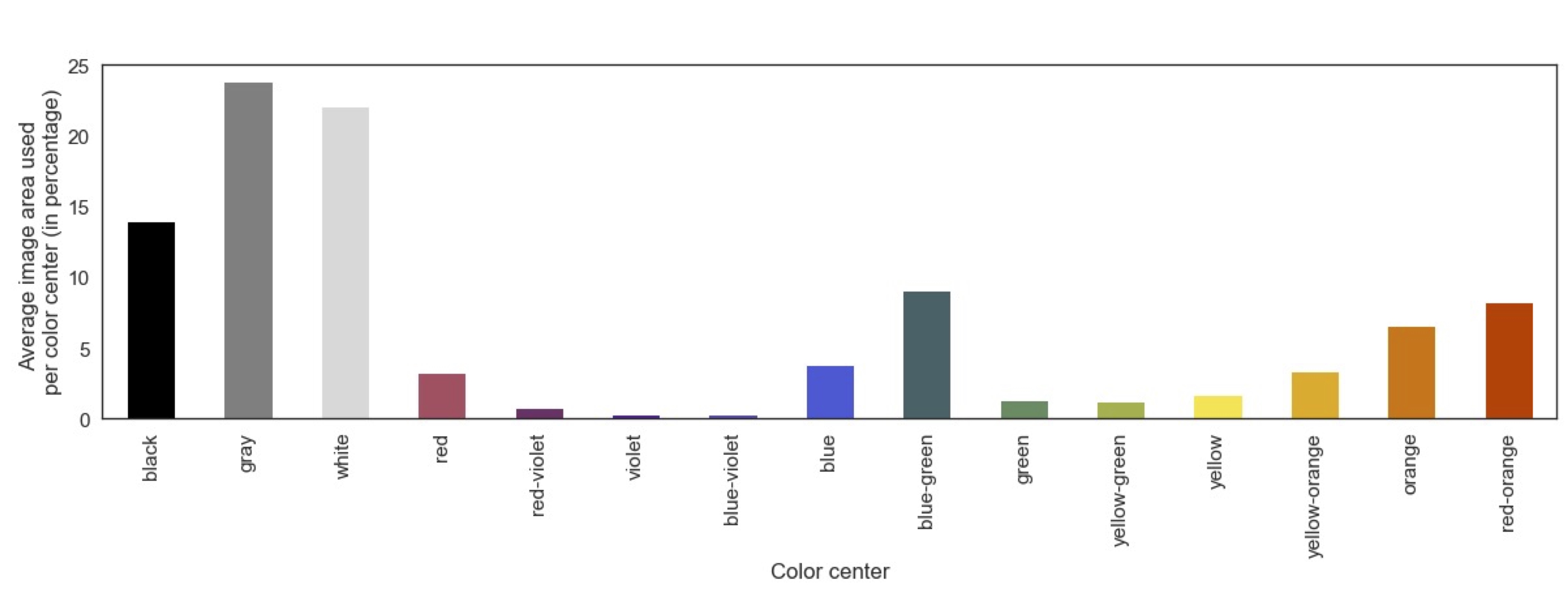}
	\caption{Extraction of color palette that makes up the COVID Museum Art (CAM) dataset}
	\label{fig:fig3}
\end{figure}

Consistent with the methodology presented, the results indicate a predominance of achromatic uses (grays, whites, and blacks), and to a lesser extent, a preference in the use of red and green-blue tones (Fig.\ref{fig:fig3}) in the CAM collection. However, it is relevant to note that the presence or absence of colors in the sample does not necessarily imply an exclusive relationship around emotions, analysis that we will discuss below.

\begin{figure}
	\centering
	\includegraphics[width=\textwidth]{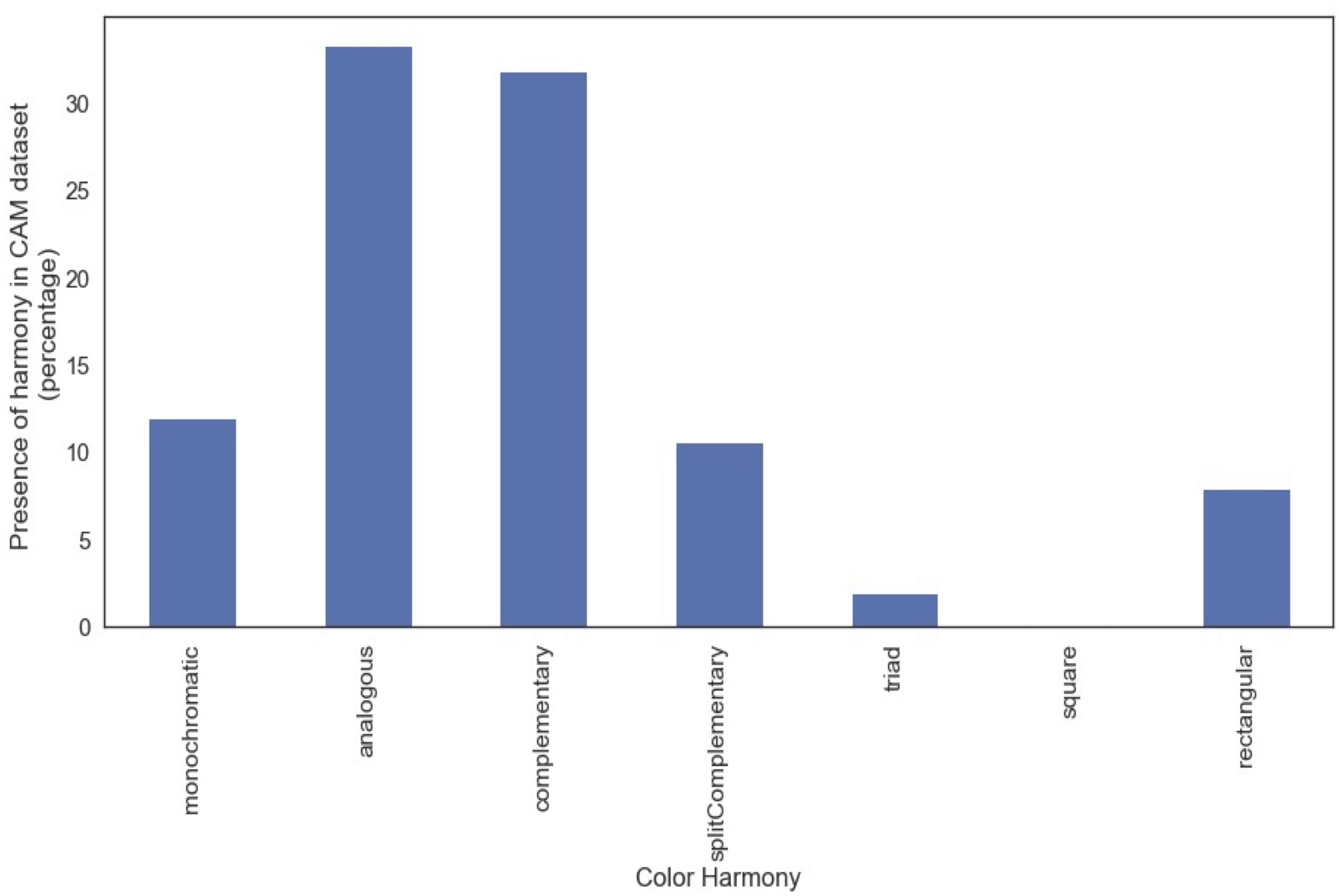}
	\caption{Frequency of color combinations present in the COVID Art Museum (CAM) dataset}
	\label{fig:fig4}
\end{figure}

In the first stage, the quantitative information of the color is done through the decomposition into HSL channels (step II). Subsequently, these results are processed, searching for possible relationships of harmonies and contrasts. An angular measurement between the colors expressed in the Itten palette is used to achieve this, described in terms of absolute frequencies. This analysis shows a more significant presence of analog and complementary combinations compared to other harmonies (Fig.\ref{fig:fig4}). Colors used in the compositions share a color with each other, or an opposite or complementary color is present in the palette.

When analyzing in detail the relationships of analog harmonies, the combinations {yellow-orange, orange, red-orange}, {orange, red-orange, red} and {yellow, yellow-orange, orange} stand out, which correspond to 72\% of the total of analog harmonies (Table~\ref{tab:table_2}). In contrast, analogous harmonies with blue, green, and violet colors account for less than 10\% of the sample. These results are relevant since the frequency of appearance of the red-orange color shows a high frequency for analogous harmonies, and the blue-green combination shows a high frequency in complementary harmonies (Table~\ref{tab:table_3}). In the case of complementary combinations, the ratio {blue-green, red-orange} is shown with a higher frequency, which corresponds to 30\% of the total of these compositions (Table~\ref{tab:table_3}).

\begin{table}
	\caption{Frequency of analogous harmonies present in the COVID Art Museum (CAM) dataset.}
	\centering
	\scalebox{1}{
	\begin{tabular}{p{9cm}rr}
Analogous Color Harmony  & Count& Percentage \\
\toprule
(yellow-orange, orange, red-orange)&	289&	33.8\% \\
(orange, red-orange, red)&	179&	20.9\%\\
(yellow, yellow-orange, orange)&	146&	17.1\%\\
(yellow-green, yellow, yellow-orange)&	62&	7.3\%\\
(red-orange, red, red-violet)&	51&	6.0\%\\
(blue-green, green, yellow-green)&  	29&	3.4\%\\
(blue, blue-green, green)&	25&	2.9\%\\
(red-violet, violet, blue-violet)&	17&	2.0\%\\
(red, red-violet, violet)&	16&	1.9\%\\
(violet, blue-violet, blue)&	15&	1.8\%\\
(green, yellow-green, yellow)&	15&	1.8\%\\
\toprule
	\end{tabular}}
	\label{tab:table_2}
\end{table}

\begin{table}
	\caption{Frequency of complementary combinations present in the COVID Art Museum (CAM) dataset}
	\centering
	\scalebox{1}{
	\begin{tabular}{p{9cm}rr}
Complementary Color Harmony  & Count& Percentage \\
\toprule
(blue-green, red-orange)&	369&	33\% \\
(blue, orange)&	187&	15\%\\
(red, green)&	30&	2\%\\
(yellow-green, red-violet)&	13&	1\%\\
(yellow-orange, blue-violet)&	9&	1\%\\

\toprule
	\end{tabular}}
	\label{tab:table_3}
\end{table}

\subsection{Emotion and color}
Are colors and harmonies associated with a certain emotion? As we have previously detected, there is a high prevalence of analog and complementary harmonies. For analogous and complementary harmonies cases, these are present in 33\% and 31\% of the works of the CAM, respectively. In the case of monochromatic, Split-complementary, and rectangular harmonies represent less than 12\% each.

\begin{figure}
	\centering
	\includegraphics[width=\textwidth]{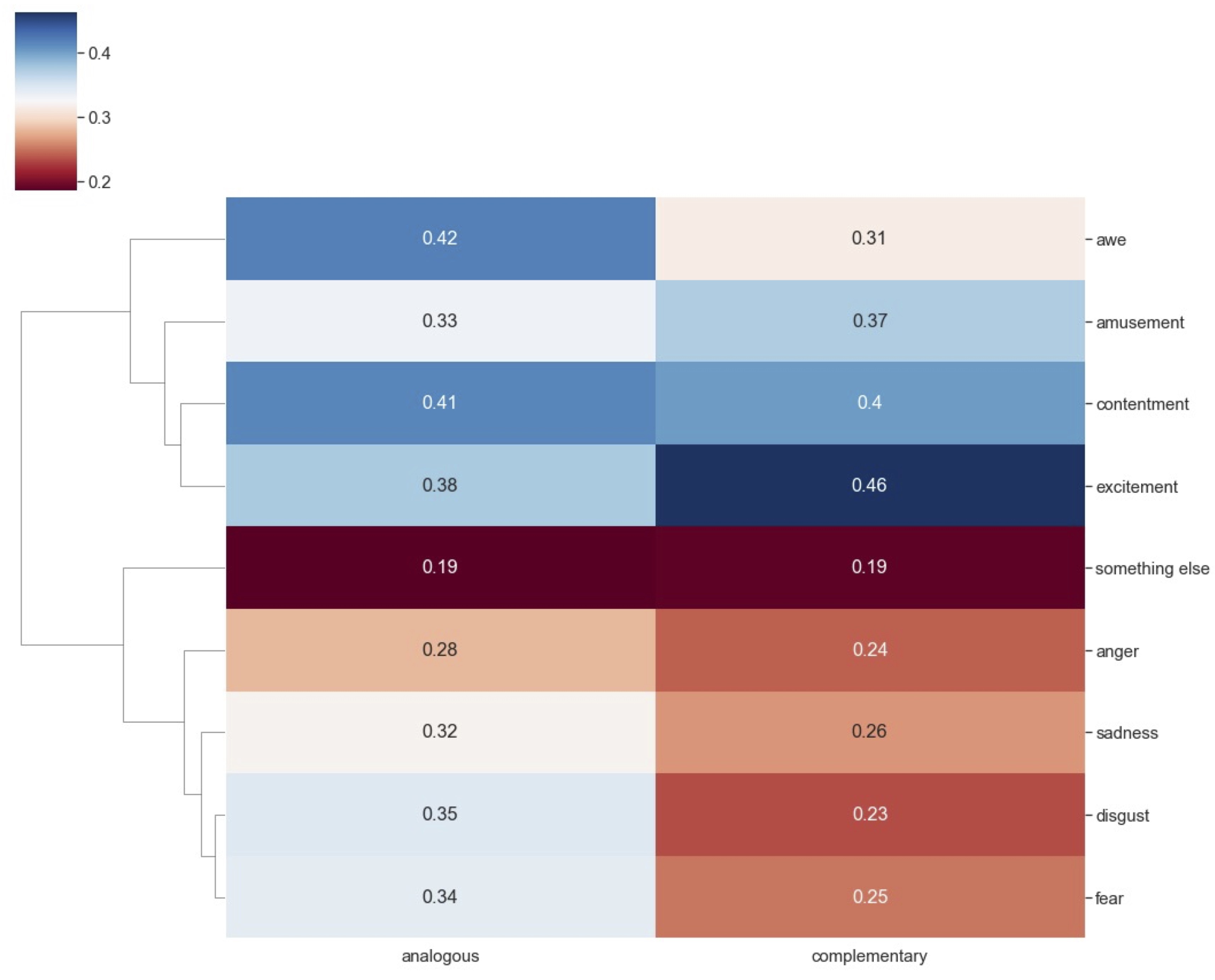}
	\caption{Correlation analysis between harmonies and emotion present in the COVID Museum Art (CAM) dataset}
	\label{fig:fig5}
\end{figure}

Experimental results indicate a slight relationship between emotion and analog harmony over complementary color combinations. In some cases, this relationship is expressed with a more significant correlation, as in the case of awe (42\%). However, in the case of complementary harmonies, excitement emotion (46\%) presents a more significant correlation versus analog harmony (see Fig.\ref{fig:fig5}).  In the case of fear and disgust emotions, it is observed that there is a greater presence of analogous versus complimentary harmonies. In the rest of the emotions, the differences between the two are minor, the emotion being inconclusive over the analog or complementary harmony.

\subsection{Correlation analysis}
\begin{figure}
	\centering
	\includegraphics[width=\textwidth]{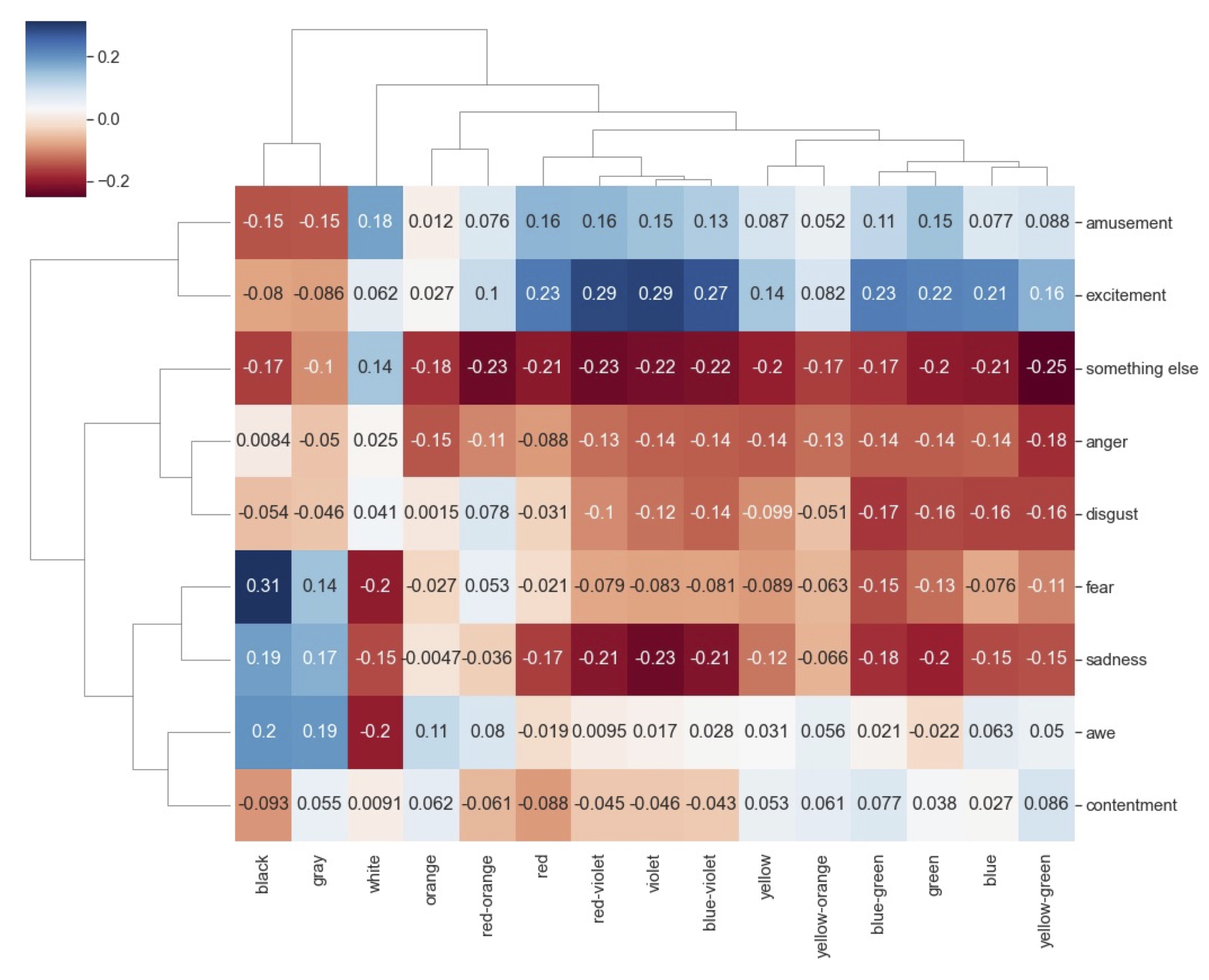}
	\caption{Correlations between each color expressed in Itten's palette and inferred emotions on the COVID Museum Art (CAM) dataset}
	\label{fig:fig6}
\end{figure}

Some relevant correlations have been found at the level of color and emotion. For example, the relationships between black and fear, red, red-violet, violet, blue-violet, and arousal have a high correlation value compared to other combinations (Fig.\ref{fig:fig6}). There is a negative correlation in the case of the emotions fear, disgust, and sadness. This means that the lower the presence of the colors violet, red-violet, blue-violet, yellow-green, blue, and green, the greater the presence of these emotions.  On the other hand, satisfaction and fear present a weak correlation with most colors, not being conclusive in any sense (greater or lesser).

We can highlight that fun is positively associated with all colors, except black and gray. This correlation varies in some colors over others, where red, violet, and green colors stand out slightly.  Finally, fear and sadness directly correlate with the colors black and gray, and inversely to white.

\subsection{A priori algorithm}
\begin{figure}
	\centering
	\includegraphics[width=\textwidth]{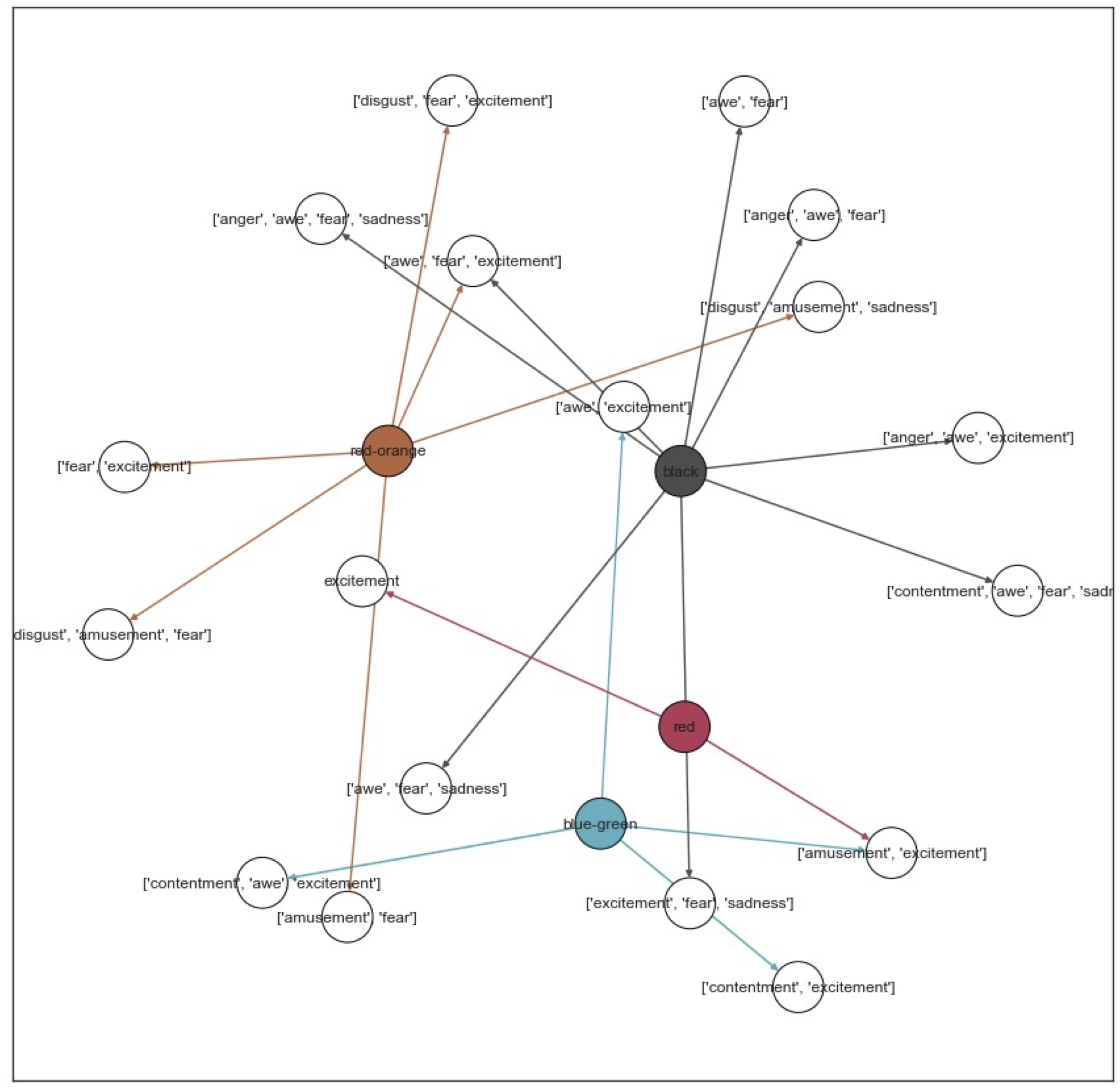}
	\caption{Relationship between color and emotion through the a priori algorithm (Agrawal et al., 1996)}
	\label{fig:fig7}
\end{figure}

Another way to analyze the relationships between colors and emotions is through the a priori algorithm \citep{agrawal_fast_1996}. This research uses the relationship {1 color} to {$n$ emotions}, where $n$ is the set of emotions detected when a specific color is observed (Fig.\ref{fig:fig7}). The relationships analyzed have a lift greater than 1. The results clearly express how some colors induce positive emotions (Blue-green), and in other cases, negative emotions (red-orange) are consistent with the correlation relationships found in Fig.\ref{fig:fig6}.

\begin{figure}
	\centering
	\includegraphics[width=\textwidth]{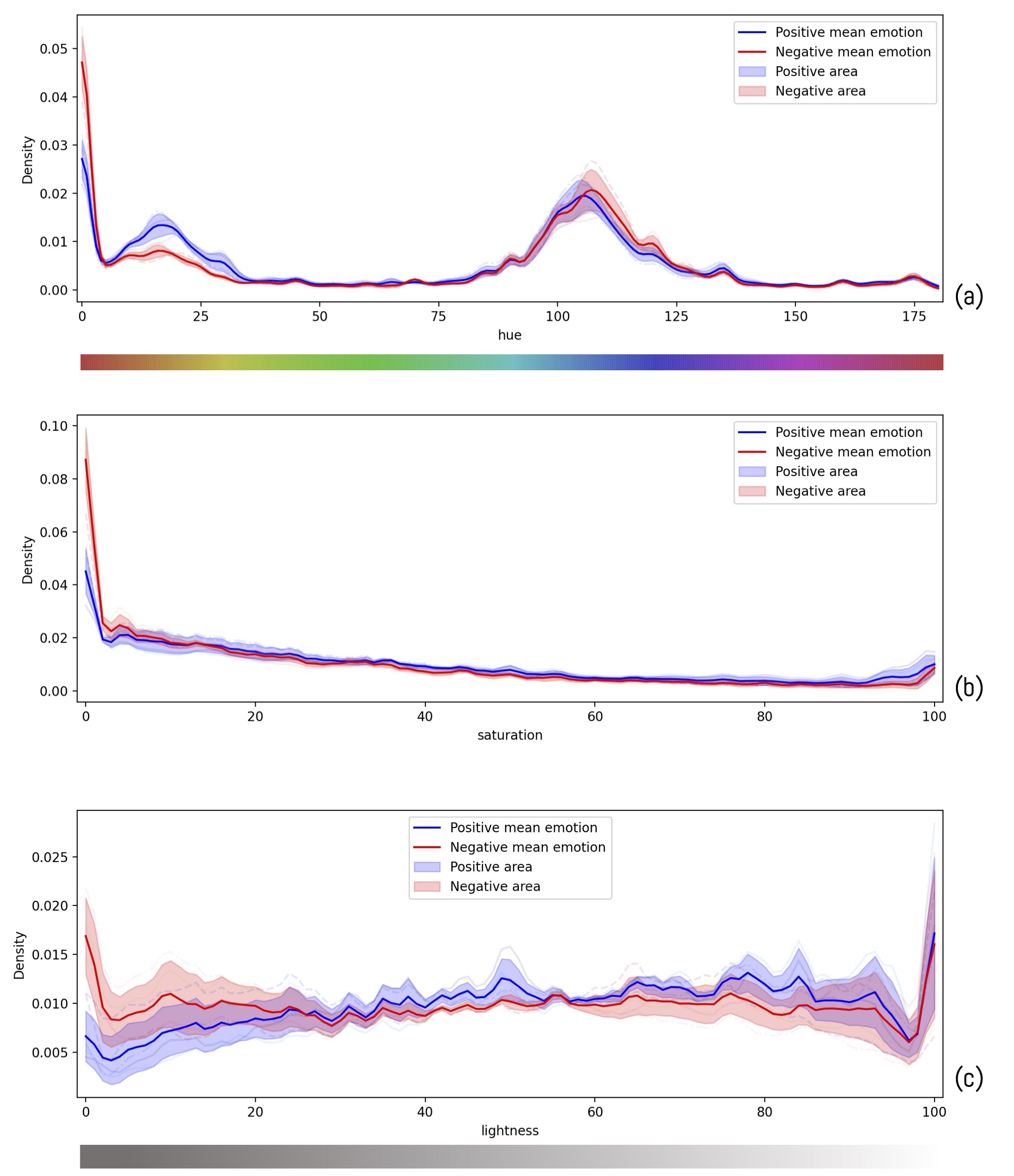}
	\caption{Emotion ratio around color, saturation and illumination (HSL).  Positive emotions={'amusement','awe','contentment','excitement'}, Negative emotions = {'anger','disgust','fear','sadness'}}
	\label{fig:fig8}
\end{figure}

At the level of the HSL transformation, some relevant differences in emotion are observed according to the Color, Lighting, and Luminance level.  For example, the emotion-arousal-fun, fear-sadness, and rage-disgust pairs exhibit similar behavior with color change (Fig.\ref{fig:fig8}a). However, the emotion category ''something else'' is the one that presents a more significant divergence compared to the other categories (Fig.\ref{fig:fig8}a). 

Regarding saturation, the main differences are in the lower value levels (<5\%) (Fig.\ref{fig:fig8}b).  In the case of the emotions of excitement-entertainment-fun, it presents a slightly greater saturation than the emotions categorized as negative in the higher saturation levels.

On the other hand, the data show that the most significant divergences are found in lighting (Fig.\ref{fig:fig8}c). There is no pattern about the emotions and lighting present in the dataset in this case. It is only possible to determine that the emotions of sadness and fear present a higher level of intensity the lower the level of light. Instead, the emotions of fun, joy,  and disgust show a higher value as the intensity of the light increases.

\section{Discussion}
According to the results obtained in the CAM collection, there is a predominance of achromatic colors that can indicate the mood of the majority if we follow the research of Carruthers \citeyear{carruthers_manchester_2010}. Although gray is predominant in the relationship between emotions and tones, it is not decisive.  If we focus on chromatic tones, the most used are blue-green and red-orange. However, yellow (as the primary color) is present in both and the following most used shades in the mixture, which causes the palette to tend towards warm color temperatures.  This is verified in the relationship of colors in analogous harmonies, where the combination yellow-orange, orange, red-orange, is the most present in the sample. The blue-green and red-orange ratio is the most numerous regarding the frequency of complementary colors.

In the relationship between these two ways of composing visually through color and emotions, the analogs show a higher score in relation to awe and a minimum superiority with contentment over complementary ones. On the other hand, the complementary ones especially highlight the excitement and slight amusement. In general, it is observed that in both compositions (analogous and complementary), the relationship with positive emotions is practically similar, and the tendency towards these emotions is shown in the sample.

Another significant result is that negative emotions (anger, sadness, disgust, and fear) are concentrated in analog compositions. In contrast, it moves away from these emotions in complementary compositions, which is logical since this type of combination creates very visually vibrant images \citep{rosenbloom_color_2006}.

The greater use of compositions through analogs, with predominance in positive emotions, can be explained by being the combination of colors closer to nature and producing more pleasure to the eyes and a more positive attitude \citep{white_color_2021}. This result demonstrates part of the conclusions of previous research \citep{cano-martinez_quantitative_2021}, where one of the most used elements of representation in the CAM collection are those related to nature. These results reveal that the pandemic, above the negative causes on health, the economy, and social damage to people, generates a response in the artistic world that is expressed as a necessity to avoid confinement and isolation  \citep{white_color_2021,cano-martinez_quantitative_2021}.

At the level of color-emotion relationship, a clear relationship is observed between black with the emotion of fear and sadness, while the colors red, red-violet, blue-violet is related to arousal and influence and giving room to the emotion of fear if its presence is reduced, along with the yellow-green and blue-green color. Fun is the emotion that most closely relates to colors.

The emotional relationship has also been analyzed depending on the attributes of the color (hue, chroma, and lighting), finding that the pairs of emotions present a similar behavior with the hue, again showing the red-orange as negative and the bluish tones is where the positive emotions are concentrated  \citep{schloss_blue_2020}, reinforcing the result with the a priori algorithm. 

Regarding saturation, although the result is very slight, it is shown that, at a higher level, positive emotions are presented, coinciding with Palmer and Schloss \citeyear{palmer_ecological_2010}, Valdez \citeyear{valdez_effects_1994}, Manav \citeyear{manav_color-emotion_2007}, Schloss \citeyear{schloss_blue_2020}, Gao \citeyear{gao_investigation_2006}, Suk \citeyear{suk_emotional_2010}. On the levels of illumination, another attribute where research agrees as a determinant in emotions in relation to color, no conclusive patterns have been found. Still, certain positive emotions are shown with the increase in luminosity.  These latest results are consistent with what has been expressed by previous research \citep{russell_distinguishing_1974}.

\section{Conclusions}
The pandemic caused by COVID-19 has meant an alteration of our daily lives, which has impacted us emotionally at different levels. The large number of images that make up the Covid Art Museum (CAM) collection on the social network Instagram becomes a valuable source of visual stories that can be analyzed to understand the emotional impact during this period at an individual and collective level.

The formal aspects and strategies used by the creators of the CAM images must be extracted and analyzed to understand this discourse. In this research, we have focused on color and its attributes, the element with the greatest capability to transmit emotions.

Given the large number of images, a methodology has been used that combines automatic processing tools, including deep learning tools for the classification of emotions and analytical tools of image processing and Machine Learning for the analysis of colors.  This methodology can be used on other datasets and offers a new type of analysis on unlabeled data sets.

In conclusion, in the sample obtained from the CAM, emotional discourses with positive tendencies prevail, which are shown by a predominance in the use of warm color palettes, the a priori algorithm, in the relationship between compositions of colors and emotions, and the attributes of color.

\section{Limitations and Future research}
This research opens up new lines of future development by analyzing color and its relationship to emotions. As presented, there are still opposing opinions on whether the color-emotion relationship is universal.  Although this research uses images created by Internet users, which can give it a global character, the geographical origin, ages, or genders of the creators of the images in the sample are unknown.  By obtaining this information, significant results could be contributed to the discussion on the color-emotion relationship and its universality.

Likewise, it could be interesting to know the intention at an emotional level of the creators of the works of the CAM collection, which could enrich and consolidate the results. 
Expanding the investigation could be done through surveys or interviews. However, the large number of images in the exhibition makes this proposal difficult.

On the other hand, as color and its relationship with emotions can be associated with objects as reviewed in the literature, this research will continue to contrast the results of a previous article \citep{cano-martinez_quantitative_2021}, where the representation elements were extracted, and with the results obtained in this research, significant progress can be made in the study of the communication of emotions in visual creation.

\bibliographystyle{unsrtnat}

\end{document}